%% file: 00.main.tex
\documentclass[conference,9pt]{IEEEtran}
\IEEEoverridecommandlockouts

\usepackage{cite}
\usepackage{amsmath,amssymb,amsfonts}
\usepackage{algorithmic}
\usepackage{graphicx}
\usepackage{textcomp}
\usepackage{xcolor}
\usepackage{url}

\usepackage{bm}
\usepackage{enumitem}
\usepackage{booktabs} 
\usepackage{amsmath}

\usepackage{multirow}
\usepackage{threeparttable}
\usepackage{makecell}
\newcommand{\cmark}{\ding{51}\xspace}
\newcommand{\xmark}{\ding{55}\xspace}

\usepackage{xspace}
\usepackage{pifont}
\usepackage{tablefootnote}

\def\BibTeX{{\rm B\kern-.05em{\sc i\kern-.025em b}\kern-.08em
    T\kern-.1667em\lower.7ex\hbox{E}\kern-.125emX}}
\begin{document}

\title{Large Language Model Can Transcribe Speech in\\Multi-Talker Scenarios with Versatile Instructions\vspace{2mm}

\thanks{*Equal Contribution}
}

{
\author{

\Large{\textit{Lingwei Meng*, Shujie Hu*, Jiawen Kang, Zhaoqing Li, Yuejiao Wang, Wenxuan Wu,}}\\
\Large{\textit{Xixin Wu, Xunying Liu, Helen Meng}}
\vspace{3mm}\\

The Chinese University of Hong Kong, Hong Kong SAR, China
 
} 
}

\maketitle

\begin{abstract}
Recent advancements in large language models (LLMs) have revolutionized various domains, bringing significant progress and new opportunities. Despite progress in speech-related tasks, LLMs have not been sufficiently explored in multi-talker scenarios. In this work, we present a pioneering effort to investigate the capability of LLMs in transcribing speech in multi-talker environments, following versatile instructions related to multi-talker automatic speech recognition (ASR), target talker ASR, and ASR based on specific talker attributes such as sex, occurrence order, language, and keyword spoken. Our approach utilizes WavLM and Whisper encoder to extract multi-faceted speech representations that are sensitive to speaker characteristics and semantic context. These representations are then fed into an LLM fine-tuned using LoRA, enabling the capabilities for speech comprehension and transcription. Comprehensive experiments reveal the promising performance of our proposed system, MT-LLM, in cocktail party scenarios, highlighting the potential of LLM to handle speech-related tasks based on user instructions in such complex settings\footnote{The code, model, and samples are available at \url{https://github.com/cuhealthybrains/MT-LLM}}.

\end{abstract}


\input{01.intro}

\input{02.method}

\input{03.experimental_settings}

\input{04.results_and_discussion}
\input{05.conclusion}

\clearpage  
\newpage

\bibliographystyle{IEEEbib}
\bibliography{mybib}

\end{document}

%% file: 01.intro.tex
\section{Introduction}

Large language models (LLMs) have experienced rapid and significant advancements recently, achieving or even surpassing human-level proficiency in numerous natural language processing (NLP) tasks \cite{OpenAI2023GPT4TR, touvron2023llama,touvron2023llama2}. These advancements have sparked interest in exploring the capabilities of LLM in multi-modal perception\cite{chen2024ntp}, including speech \cite{hu2024wavllm,tang2024salmonn,Qwen2-Audio}, vision \cite{OpenAI2023GPT4TR,huang2023language,peng2024grounding}, and content generation \cite{pan2024kosmosg, brooks2024video, meng2024melle}. Several studies have investigated speech-related LLMs, which typically involve a fine-tuned text LLM following speech-related instructions and pairing with auxiliary audio encoders \cite{hu2024wavllm,tang2024salmonn,Qwen2-Audio}. The audio encoder extracts acoustic representations and adapts them to the input feature space of LLM, allowing LLM to perform various speech tasks such as automatic speech recognition (ASR), speech translation (ST), speaker verification (SV), and speech question answering (SQA), among others. 
However, despite the progress, the potential of speech LLMs in cocktail party scenarios—where multiple talkers speak simultaneously and overlapping occurs—has not yet been sufficiently exploited. 

In recent year, various end-to-end approaches have garnered interest and been developed to tackle multi-talker ASR task, which involves simultaneously transcribing speech from multiple talkers. These studies are based on Permutation Invariant Training (PIT) \cite{zhang2020pit,chang2020pit,meng23sidecaricassp}, Heuristic Error Assignment Training (HEAT) \cite{lu2021surt,raj2023surt2}, or Serialized Output Training (SOT) \cite{kanda2020sot,kanda22tsot,shi24sot,kang2024disentangling,zheng2024sot} to match predictions with corresponding target labels for loss calculation. However, these approaches typically transcribe speech from all talkers indiscriminately and fail to associate transcriptions with specific talkers, unless an additional external \cite{huang2023pit_tse,masumura23joint} or internal \cite{kanda21sasot,conformer-ts-asr,masumura24_interspeech} model is employed to extract speaker information. Although several studies \cite{meng23sidecarinterspeech,meng24whisperovlp} proposed handling multi-talker ASR in conjunction with other tasks within a single model, the addressed tasks remain constrained and lack the flexibility to address various user requirements specifying talker attributes such as \textit{please transcribe the talker who said the word “strawberry”}. 

Nevertheless, the rise of large language models illuminates new possibilities for tackling such problems with a unified model. 
In this work, we leverage the powerful comprehension and instruction-following capabilities of LLM to perform speech recognition based on various instructions in multi-talker scenarios. Specifically, we utilize Llama 2 \cite{touvron2023llama2} as our foundational LLM, coupled with the Whisper \cite{radford2023whisper} encoder to extract semantic context, and WavLM \cite{chen2022wavlm} multi-layer features to capture acoustic information indicating speaker characteristics, referring to WavLLM \cite{hu2024wavllm} and SALMONN \cite{tang2024salmonn}. Corresponding adapters are designed to project audio embeddings into the LLM's input space. We denote the proposed model as MT-LLM (Multi-Talker LLM). Versatile instructions are used to prompt MT-LLM to perform tasks including (i) simultaneously transcribing the speech of multiple talkers into text, (ii) transcribing a target talker's speech given a reference audio clip, (iii) transcribing speech based on the talker's specific sex, (iv) transcribing the speech of a specified talker according to their occurrence order, (v) transcribing the speech of the talker where a given keyword appears, and (vi) transcribe the talker who speaks the specific language. The comprehensive experiments demonstrate that MT-LLM can effectively meet user's diverse requirements for transcribing multiple talkers based on instructions specifying talker attributes.
Our major contributions are threefold:
\begin{itemize}[leftmargin=1em]
    \item  We propose a pioneering effort to explore instruction-based speech recognition in multi-talker scenarios, leveraging the powerful comprehension and generation capabilities of LLM;
    \item Beyond multi-talker ASR, MT-LLM can transcribe speech from specific talkers according to six versatile instructions, demonstrating promising performances;
    \item We reveal that speech LLMs can support a more natural and effective human-computer interaction paradigm in complex speech environments, with parameter-efficient training.
\end{itemize}

%% file: 02.method.tex
\begin{figure*}[h]
    \centering
    \includegraphics[width=\textwidth]{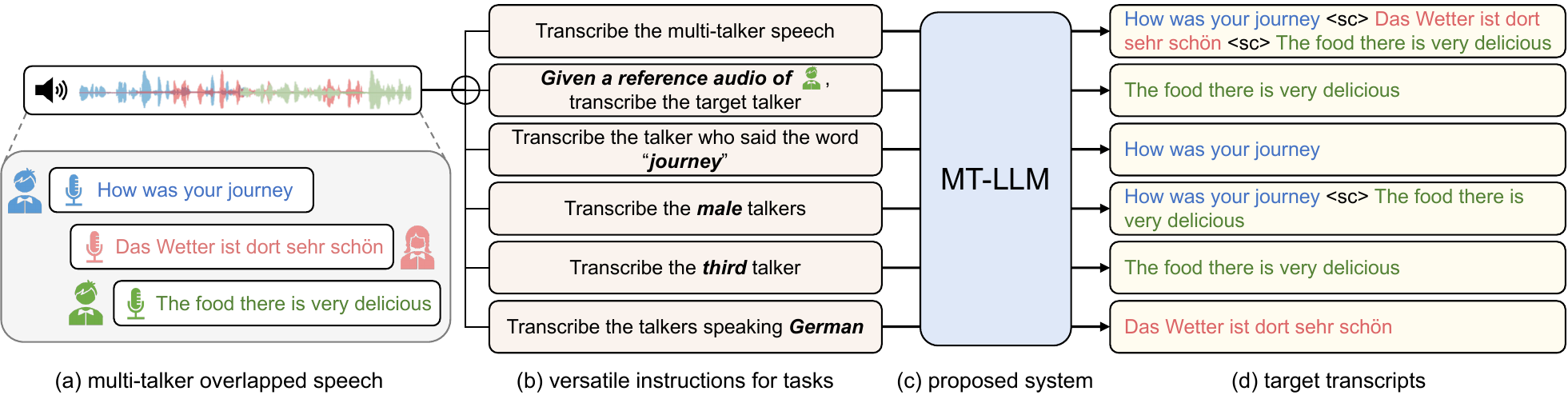}
    \caption{
       MT-LLM supports versatile ASR instructions in multi-talker scenarios. Given a multi-talker overlapped speech input (a) and a text instruction prompt (b), the proposed MT-LLM (c)
       is expected to autoregressively generate corresponding target transcripts (d). For tasks that involve multiple talkers, MT-LLM follows the SOT-style output, transcribing the utterances of multiple talkers in the order of their start times, separated by “$<$sc$>$” indicating “speaker change”.
    }
    \vspace{-1mm}
    \label{fig:overview}
\end{figure*}

\section{Methods}
We propose empowering a text-based LLM to act as a versatile instruction follower for speech recognition in multi-talker speech scenarios. The proposed method consists of three major components: a large language model as the foundational model fine-tuned with low-rank adaptation (LoRA) \cite{hu2022lora}, dual speech encoders with corresponding adapters, and training data construction. We denote the proposed model as MT-LLM in the subsequent sections.
\subsection{Problem Formulation}
This study regards the speech recognition in multi-talker scenarios as the next-token-prediction language modeling task. Conditioned on the input speech waveform $\bm{X}$ and text instruction $\bm{I}$, MT-LLM is optimized to autoregressively generate the target text output $\bm{Y}=[\bm{y}_0, \bm{y}_1, ..., \bm{y}_{N-1},]$, by maximizing the following distribution:
\begin{align}
   p(\bm{Y} \mid \bm{X}, \bm{I} ; \bm{\theta}) &= \prod_{t=0}^{T-1} p(\bm{y}_t \mid \bm{X}, \bm{I}, \bm{Y}_{<t} ; \bm{\theta})
\end{align} 
where $\bm{\theta}$ represents the parameters of MT-LLM. 

Among $\bm{Y}$, The transcripts for multiple talkers require a permutation assignment to determine the talker order, thereby addressing the label ambiguity issue \cite{yu2017pit}. Drawing on previous experiences \cite{kanda2020sot, shi24sot}, we employed the straightforward Serialized Output Training (SOT) method to address this issue. As illustrated in Fig. \ref{fig:overview} (d), SOT arranges the transcripts based on the speaking order of the talkers, with plain text “$<$sc$>$” inserted between them to signify speaker changes.

\subsection{Model Architecture}
\begin{figure}[t]
    \centering
    \includegraphics[width=0.95\columnwidth]{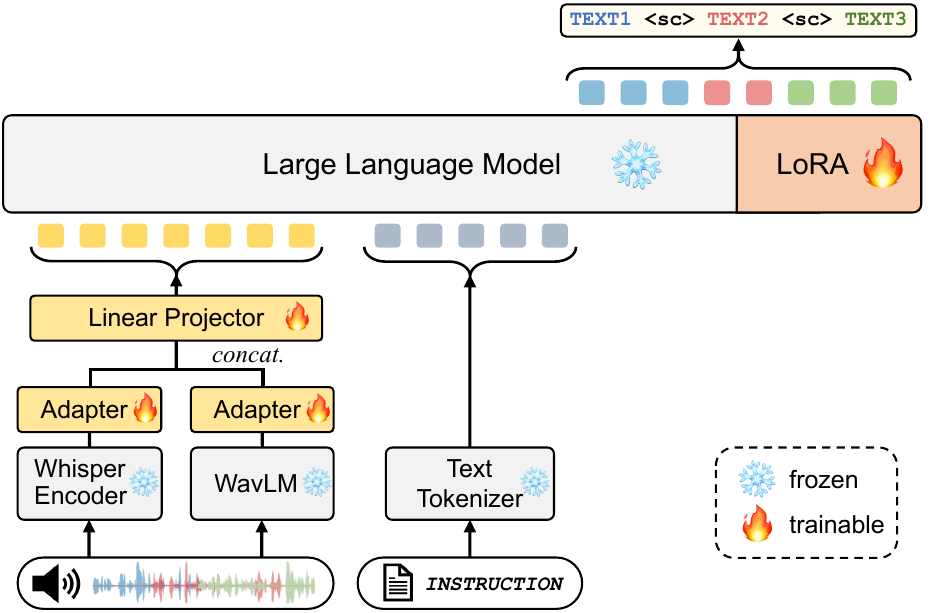}
    \caption{
      Model architecture of MT-LLM. Multi-faceted speech representations are extracted using dual speech encoders and projected into the LLM feature space. Fine-tuned with LoRA, the LLM acquires the capability to comprehend and transcribe speech in multi-talker scenarios based on text instructions.
    }
    \label{fig:framework}
\end{figure}

As shown in Fig. \ref{fig:framework}, the speech representations synthesized by the dual speech encoders are fused and subsequently projected into the feature space of the backbone LLM. The LLM then leverages both the speech input and text instructions to predict the target transcripts.  The backbone LLM and speech encoders remain frozen, while the system is fine-tuned using parameter-efficient adapters, equipping it with the capability for speech processing and comprehension.

\noindent \textbf{Dual Speech Encoders and Corresponding Adapters\quad} Referring to \cite{tang2024salmonn, hu2024wavllm}, we utilize two pre-trained speech encoders, Whisper encoder and WavLM, to capture multi-faceted speech information. Whisper \cite{radford2023whisper} is a speech recognition and translation model trained on web-scale weakly supervised data, with its encoder being sensitive to speech semantic context \cite{gong23whisperat}. We exploit its last-layer output embedding to capture rich semantic information.  In contrast, WavLM \cite{chen2022wavlm} is a self-supervised learning model trained using a masked speech prediction approach, leading to different layers encoding acoustic information sensitive to various downstream speech tasks. To better leverage the multi-scale acoustic information extracted by WavLM, we aggregate its multi-layer hidden states by summing them with learnable weights for each layer. The modality adapters for the two speech encoders, along with a linear projector, are designed to better align speech representations with the LLM feature space.

\noindent \textbf{Backbone LLM and LoRA\quad}
We exploit Llama-2-7b-chat\footnote{\url{https://llama.meta.com/llama2}}, developed by Meta AI\cite{touvron2023llama2} on extensive and diverse training data, as our foundation backbone. Llama 2 excels across a broad spectrum of NLP tasks, showcasing superior capabilities in context understanding and text generation. To incorporate the speech modality into the LLM, we employ a parameter-efficient fine-tuning technique, LoRA \cite{hu2022lora}. LoRA is applied to the key, query, value, and output weight matrices within Llama’s attention modules, enabling the model’s capability to process and comprehend speech modality inputs.

\subsection{Task Descriptions}
\label{sec:tasks}
To validate the performance of MT-LLM in executing speech recognition based on instructions in multi-talker scenarios, we simulated multi-talker overlapped speech audios from single-talker speech corpora, and designed six tasks along with corresponding instructions and generated the respective target text label. 
As illustrated in Fig. \ref{fig:overview} (b), the text instructions pertain to the following tasks:

\noindent \textbf{Multi-Talker (MT) ASR\quad} Simultaneously transcribing the speech of multiple talkers into text. This basic task challenges the model's ability to handle overlaps and distinguish between different talkers' voices within a single audio stream.

\noindent \textbf{Target-Talker (TT) ASR\quad} A random talker is selected as the target, and a 3-second audio clip of the target talker, along with 3 seconds of silence, is concatenated with the input multi-talker speech. The model is instructed to transcribe only the target talker's speech from the overlapping audio. This task tests the model's capability to differentiate and isolate the speech of a specified individual given a reference audio clip as the clue.

\noindent \textbf{Keyword-Tracing (KT) ASR\quad} For each multi-talker sample, we first collect a set of unique words that appear only once across all talkers' speech content, each with a minimum length of six characters. From this set, we randomly select a keyword and instruct the model to transcribe the speech of the talker who said that word. This task assesses the model's proficiency in tracking specific lexical items and attributing them to the correct speaker.

\noindent \textbf{Sex-Specific (SS) ASR\quad} We randomly instruct the model to transcribe all male or all female talkers from the multi-talker speech input. 
This task evaluates the model's ability to distinguish voices based on sex-related characteristics and filter the transcription accordingly.

\noindent \textbf{Order-Specific (OS) ASR\quad} We randomly instruct the model to transcribe a talker based on their appearance order. This requires the model to keep track of the sequence of speakers and accurately extract the speech of a designated individual in the order they spoke.

\noindent \textbf{Target-Lingual (TL) ASR\quad} We randomly instruct the model to transcribe the speech of talkers speaking either English or German. This task evaluates the model's ability to discriminate languages and transcribe spoken content based on the specified language.

%% file: 03.experimental_settings.tex
\section{Experimental Setup}

\subsection{Training Data Construction}
We simulated multi-talker speech audios from single-talker speech corpora to accommodate the versatile task instructions, following the protocol outlined in \cite{kanda2020sot}. Specifically, the utterances are simulated primarily from the 960-hour LibriSpeech \cite{librispeech} training set, comprising mixtures of two or three talkers. The start time for each talker is randomly sampled, resulting in overlapped mixtures.
Additionally, the 180-hour German subset of CoVoST 2\footnote{Note that CoVoST 2 is originally designed for speech translation, while we use its German subset for ASR in our study.} \cite{wang21CoVoST2}, along with its corresponding German target text, is employed to mix with LibriSpeech utterances to support the Target-Lingual ASR task.

Combined with original single-talker corpora, the total training speech data amounts to $\sim$6.3K hours, with $\sim$10\% containing German.

\subsection{Evaluation and Metrics}
We evaluate the performance of MT-LLM on versatile instruction-based tasks, using the official 2- and 3-speaker LibriSpeechMix test set and an additional home-made En-De-Mixed test set for Target-Lingual ASR task, ensuring that none of the speakers from evaluation sets are included in the training data. We also test single-talker ASR performance on LibriSpeech test-clean set and CoVoST 2 German (De) test set. 


The word error rate (WER) is calculated between the predicted text and the target labels. For speech recognition tasks involving multiple talkers, permutations with minimum errors are used to compute WER, following previous studies \cite{shi24sot, jiawen_icassp, cornell2024chime8}.

\subsection{Model Settings and Training details}
The proposed MT-LLM employs the encoder of Whisper-large-v2\footnote{https://huggingface.co/openai/whisper-large-v2} and WavLM-base\footnote{https://huggingface.co/microsoft/wavlm-base} to extract speech representations, with Llama-2-chat-7b as the backbone LLM. All parameters of above models are frozen. The LLM is fine-tuned using LoRA with a rank of 32. The adapters for both speech encoders consist of two convolution layers to down-sample and align the representations within the temporal domain, followed by a bottleneck adapter \cite{houlsby2019adapter} and a linear layer. The outputs of both adapters have a time stride of 80 ms and a dimension of 2,048. The total number of parameters in MT-LLM is 7.55 billion, of which 1\% (76.6 million) are trainable.

MT-LLM is trained on 32 NVIDIA A100-40G GPUs with a batch size equivalent to 60 seconds per GPU for 150K updates. We optimize the model using AdamW optimizer, warming up the learning rate to a peak of 1e-4 over the first 10\% updates, followed by a linear decay. All tasks are mixed together for training to develop a unified model.

%% file: 04.results_and_discussion.tex
\input{tables/01.mt-asr}
\section{Results and Discussion}

\subsection{Results on Multi-Talker ASR}
\label{sec:mtasr}
We evaluate the single-talker ASR performance on the LibriSpeech and CoVoST 2 German test set, and multi-talker ASR performance on the 2- and 3-speaker LibriSpeechMix and the home-made En-De-Mixed test set. 
As shown in Table \ref{tab:mtasr}, the proposed MT-LLM demonstrates promising results across all datasets, outperforming D2V-Sidecar-DB \cite{meng23sidecarinterspeech}, a recent representative work employing PIT. 
SOT-Conformer \cite{kanda21sasot} is a state-of-the-art model specifically designed and extensively trained for multi-talker ASR on the LibriSpeechMix dataset. On the 2-speaker LibriSpeechMix, MT-LLM achieves results comparable to SOT-Conformer, while lags behind on the 3-speaker set. Given that MT-LLM serves as an exploratory work aimed at exploring the potential of LLM for versatile instruction-based ASR in complex environments, rather than solely pushing the limits of multi-talker ASR, we argue that the gap is understandable. 
Despite being trained on both single-talker and multi-talker data, we note that SOT-Conformer falls short in single-talker scenarios, indicating its specialization in multi-talker ASR impairs its single-talker performance. In contrast, MT-LLM performs consistently good in single-talker ASR, demonstrating that its instruction-understanding capabilities help maintain high performance in simpler, single-talker settings. SALMONN \cite{tang2024salmonn} is a speech LLM trained on various speech-related tasks including overlapped speech recognition. However, using its official prompt, we observed significant inferior performance compared to our approach as shown in Table \ref{tab:mtasr}.

We list wav2vec2-Large-XLSR-53-German model\footnote{\url{https://huggingface.co/jonatasgrosman/wav2vec2-large-xlsr-53-german}} performance on De for reference. Given that the audio samples containing German is relatively limited and noisier than LibriSpeech, MT-LLM still demonstrates satisfactory results on De and En-De-Mix test sets, showcasing the model's ability to simultaneously distinguish and transcribe multiple languages from multi-talker speech.

\subsection{Results on Versatile Tasks Based on Instructions}
We investigate the performance of MT-LLM on various ASR tasks described in Section \ref{sec:tasks}. The evaluations are conducted using the 2- and 3-speaker LibriSpeechMix and En-De-Mix test sets, with the results presented in Table \ref{tab:instruction2spk} and Table \ref{tab:instruction3spk}.

To validate MT-LLM's ability to accurately capture and transcribe a specified talker in multi-talker scenarios, we evaluate a \textit{best-matching} result as a reference. Specifically, for each sample, we select the target text of the correct talker based on instructions specifying talker attributes (such as sex or keywords), and calculate WER against each ASR transcript of multiple talkers produced by MT-LLM in Section \ref{sec:mtasr}. The lowest WER obtained is reported as the best-matching result, eliminating the impact of speaker confusion. This result is compared with the model's performance when directly executing instruction-based ASR tasks, thereby reflecting MT-LLM's capability to follow instructions and correctly identify the intended talker.

The results in Table \ref{tab:instruction2spk} and Table \ref{tab:instruction3spk} highlight the effectiveness of the MT-LLM model across different instruction-based ASR tasks. In both the 2-speaker and 3-speaker mixed scenarios. MT-LLM achieves impressive performance for Target-Talker (TT), Keyword Tracing (KT), and Sex-Specific (SS) ASR tasks, which are within a reasonable range of the best-matching results. This suggests that MT-LLM is proficient at isolating and accurately transcribing speech from specified talkers based on instructions focusing on different speaker attributes. 
However, for Order-Specific (OS) ASR tasks, there is a gap compared to the best-matching result, indicating that MT-LLM slightly falls short in determining the order in which the talker appears. We anticipate significant improvements by using additional positional embedding for the speech part, employing larger speech modality adapters, or fully fine-tuning the speech encoder with heavier training as in \cite{kanda21sasot}. 
Target-Lingual (TL) ASR task presents a more pronounced challenge for the model, due to that the mixed audios containing German are relatively limited and noisier compared to English set. We observe that performance of the TL task in both English and German is close, resulting in similar WER to best-matching situation. This indicates room for improvement through language-specific embeddings or higher-quality multi-lingual data.

For the more complex 3-speaker scenarios, MT-LLM consistently manages to handle the added intricacies of increased overlapping of multiple talkers, despite a rise in WER compared to the 2-speaker cases. This discrepancy illustrates the challenge posed by denser overlapping in speech, yet MT-LLM still demonstrates a competent ability to follow instructions and transcribe the target speaker accordingly.

\subsection{Ablation Study}
First, we conduct an ablation study to examine the impact of using dual speech encoders. We train two models specifically on the multi-talker (MT) ASR task: one using dual speech encoders and the other using only the Whisper encoder. As shown in Table \ref{tab:ablation}, incorporating the WavLM encoder into the architecture leads to notable performance improvements, underscoring the importance of multi-layer acoustic information captured by WavLM in improving speech recognition accuracy, particularly in three-talker scenarios. 

We also investigate the effect of multi-task training on improving the basic MT ASR task. As illustrated in Table \ref{tab:ablation}, training on a variety of tasks not only endows the system with the versatility to handle diverse ASR tasks but also improves performance on the basic MT ASR task. This indicates that the different tasks are interdependent and complementary, helping to supervise the model and enhance its overall speech comprehension capabilities.


\subsection{Limitations and Future Work}
Despite MT-LLM's promising performance on various ASR tasks in multi-talker scenarios, we acknowledge several limitations. First, the primary intention of this study is to explore the ability of LLMs to capture specific talkers according to instructions, and transcribe their speech in multi-talker scenarios. Therefore, MT-LLM is not designed to be a universal speech LLM for a broader spectrum of tasks. Second, the experiments are conducted on limited simulated datasets for insightful observation, rather than on real-world datasets, due to resource constraints. 
In the future, we anticipate the development of a more comprehensive speech LLM for cocktail party environments with meticulously crafted training data and schemes.

\input{tables/02.instruction-2spk}
\input{tables/03.instruction-3spk}
\input{tables/04.ablation}

%% file: tables/01.mt-asr.tex
\begin{table}[h]


  \caption{Single-talker and multi-talker ASR performance on English and German test set. Evaluated by WER (\%).}
  \label{tab:mtasr}
  \centering 
  \setlength{\tabcolsep}{0.8mm}
  {
  \begin{tabular}{lcccccc}

    \toprule
  \multirow{2}{*}[-0.5ex]{\textbf{System}} & \textbf{LibriSpeech} & \multicolumn{2}{c}{\textbf{LibriSpeechMix}} & \textbf{De} & \multicolumn{2}{c}{\textbf{En-De-Mix}} \\
 \cmidrule(lr){2-2} \cmidrule{3-4} \cmidrule(lr){5-5} \cmidrule{6-7}
    &  1-spk  &  2-spk & 3-spk &  1-spk & 2-spk & 3-spk     \\
  \midrule
    D2V-Sidecar-DB \cite{meng23sidecarinterspeech} & -  & 7.5 & 11.9&-&-&-\\
   SOT-Conformer \cite{kanda21sasot}&  3.6 & 4.9  & 6.2 &-&-&- \\ 

    SALMONN-7B \cite{tang2024salmonn} & 2.4 & 32.9 & 45.9 & - & - & -\\
    XLSR-Large-De \cite{xlsr} & -& - & - & 12.1 & - & - \\
    \midrule
    MT-LLM (ours) &  2.3  &  5.2 & 10.2  & 11.7 & 21.0  & 22.8 \\
\bottomrule
\end{tabular}
}

\end{table}

%% file: tables/02.instruction-2spk.tex
\begin{table}[t]

\caption{Results for versatile ASR tasks on 2-speaker LibriSpeechMix and En-De-Mix test sets. Evaluated by WER (\%).}
\vspace{-1mm}
\label{tab:instruction2spk}
\centering 

\setlength{\tabcolsep}{2.5mm}
{
\begin{tabular}{lccccc}

\toprule
\multirow{2}{*}[-0.5ex]{\textbf{System}} & \multicolumn{4}{c}{\textbf{LibriSpeechMix 2-spk}} & \textbf{En-De-Mix 2-spk} \\
\cmidrule(lr){2-5} \cmidrule(l){6-6}
& \textbf{TT} & \textbf{KT} & \textbf{SS} & \textbf{OS} & \textbf{TL}   \\
\midrule
best-matching  & 5.5 & 4.8 & 4.9 & 5.7 & 21.0\\
MT-LLM & 6.7 & 5.0 &  5.5 & 9.0 &  21.8 \\

\bottomrule
\end{tabular}

}
\vspace{-2mm}
\end{table}

%% file: tables/03.instruction-3spk.tex
\begin{table}[t]

\caption{Results for versatile ASR tasks on 3-speaker LibriSpeechMix and En-De-Mix test sets. Evaluated by WER (\%).}
\vspace{-1mm}
\label{tab:instruction3spk}
\centering 

{
\begin{tabular}{lccccc}

\toprule
\multirow{2}{*}[-0.5ex]{\textbf{System}} & \multicolumn{4}{c}{\textbf{LibriSpeechMix 3-spk}} & \textbf{En-De-Mix 3-spk} \\
\cmidrule(lr){2-5} \cmidrule(l){6-6}
& \textbf{TT} & \textbf{KT} & \textbf{SS} & \textbf{OS} & \textbf{TL}   \\
\midrule
best-matching & 12.0 & 9.7 & 12.4 & 11.7 & 24.0 \\
MT-LLM & 16.2 & 12.6 & 15.0 & 15.4 & 24.1  \\

\bottomrule
\end{tabular}

}
\vspace{-2mm}

\end{table}

%% file: tables/04.ablation.tex
\begin{table}[!t]
\caption{Ablation study on the usage of dual speech encoders and multi-task training. Evaluated by WER (\%).}
\vspace{-1mm}
\label{tab:ablation}
\centering 

\setlength{\tabcolsep}{3.2mm}

{
\begin{tabular}{ccccccc}

\toprule
\multirow{2}{*}[-0.5ex]{\textbf{\makecell{With\\WavLM}}} & \multirow{2}{*}[-0.5ex]{\textbf{\makecell{Multi-Task\\Training}}}   & \textbf{LibriSpeech} & \multicolumn{2}{c}{\textbf{LibriSpeechMix}}  \\
\cmidrule(lr){3-3} \cmidrule(l){4-5} 
&  & 1-spk  &  2-spk & 3-spk    \\
\midrule

\xmark & \xmark& 4.3  &  6.6 & 16.1    \\
\cmark & \xmark &  2.4  &  5.5 & 10.7    \\
\cmark & \cmark & 2.3  &  5.2 & 10.2    \\
\bottomrule
\end{tabular}

}


\vspace{-2mm}
\end{table}

%% file: 05.conclusion.tex
\section{Conclusion}
In this work, we present a pioneering exploration into the use of large language models (LLMs) for instruction-based speech recognition in multi-talker scenarios. We utilize the Whisper encoder to extract semantic context information and WavLM to capture multi-layer acoustic information indicating speaker characteristics, thereby enabling the foundational LLM to effectively handle speech modality input.
With parameter-efficient fine-tuning, the proposed MT-LLM demonstrates remarkable capabilities in comprehending and transcribing speech based on versatile instructions related to multi-talker ASR, target talker ASR, and ASR based on specific talker attributes such as sex, occurrence order, language, and keyword spoken. Comprehensive experiments reveal promising performance in complex multi-talker environments, highlighting the potential of LLMs to enhance speech-related tasks and improve human-computer interaction in challenging settings.

\section{Acknowledgments}
This research is partially supported by the HKSARG Research Grants Council’s Theme-based Research Grant Scheme (Project No. T45-407/19N) and by the CUHK Stanley Ho Big Data Decision Analytics Research Centre.